\documentclass[runningheads, envcountsame, a4paper]{llncs}

\usepackage{amsmath,amsfonts,amssymb}
\usepackage[ruled,linesnumbered,boxed]{algorithm2e}
\usepackage{mwe,graphicx,tabu,subcaption,float}
\usepackage[table,x11names]{xcolor}
\usepackage{color,soul}
\usepackage{tcolorbox}
\usepackage{hyperref}
\usepackage{color}

\ifdefined\paperDraft
 \def\mycomment#1{{\color{blue}[\textit{#1}]}}
 
 \else
 \def\mycomment#1{}

\fi

\SetKwComment{Comment}{$\triangleright$\ }{}

\SetKwInOut{Parameter}{Parameters}
\begin{document}

\title{RaKUn: Rank-based Keyword extraction via Unsupervised learning and Meta vertex aggregation}
\titlerunning{RaKUn}

\author{Bla\v{z} \v{S}krlj\inst{1,2} \and
Andra\v{z} Repar\inst{1,2} \and
Senja Pollak\inst{2,3}}
\authorrunning{\v{S}krlj, Repar and Pollak.}
%
\institute{Jo\v{z}ef Stefan International Postgraduate School \and
Jo\v{z}ef Stefan Institute, Slovenia \and 
Usher Institute, Medical School, University of Edinburgh, Edinburgh\\
\email{\{blaz.skrlj,senja.pollak@ijs.si\}, repar.andraz@gmail.com}}
\maketitle              
\begin{tcolorbox}
 The final authenticated publication is available online at \url{10.1007/978-3-030-31372-2_26}.
 \end{tcolorbox}
\begin{abstract}

Keyword extraction is used for summarizing the content of a document and supports efficient document retrieval, and is as such an indispensable part of modern text-based systems. We explore how load centrality, a graph-theoretic measure applied to graphs derived from a given text can be used to efficiently identify and rank keywords. Introducing meta vertices (aggregates of existing vertices) and systematic redundancy filters, the proposed method performs on par with state-of-the-art for the keyword extraction task on 14 diverse datasets. The proposed method is unsupervised, interpretable and can also be used for document visualization.

\end{abstract}


\section{Introduction and related work}

Keywords are terms (i.e. expressions) that best describe the subject of a document \cite{beliga2015overview}. A good keyword effectively summarizes the content of the document and allows it to be efficiently retrieved when needed. Traditionally, keyword assignment was a manual task, but with the emergence of large amounts of textual data, automatic keyword extraction methods have become indispensable. Despite a considerable effort from the research community, state-of-the-art keyword extraction algorithms leave much to be desired and their performance is still lower than on many other core NLP tasks \cite{hasan2014automatic}.
The first keyword extraction methods mostly followed a supervised approach \cite{hulth2003improved,nguyen2010wingnus,witten2005kea}: they first extract keyword features and then train a classifier on a gold standard dataset. For example, KEA \cite{witten2005kea}, a state of the art supervised keyword extraction algorithm is based on the Naive Bayes machine learning algorithm. While these methods offer quite good performance, they rely on an annotated gold standard dataset and require a (relatively) long training process. In contrast, unsupervised approaches need no training and can be applied directly without relying on a gold standard document collection. They can be further divided into statistical and graph-based methods. The former, such as YAKE \cite{campos2018yake,campos2}, KP-MINER \cite{el2009kp} and RAKE \cite{rose2010automatic}, use statistical characteristics of the texts to capture keywords, while the latter, such as Topic Rank \cite{bougouin2013topicrank}, TextRank \cite{mihalcea2004textrank}, Topical PageRank \cite{sterckx2015topical} and Single Rank \cite{wan2008single}, build graphs to rank words based on their position in the graph. Among statistical approaches, the state-of-the-art keyword extraction algorithm is YAKE \cite{campos2018yake,campos2}, which is also one of the best performing keyword extraction algorithms overall; it defines a set of five features capturing keyword characteristics which are heuristically combined to assign a single score to every keyword. On the other hand, among graph-based approaches, Topic Rank \cite{bougouin2013topicrank} can be considered state-of-the-art; candidate keywords are clustered into topics and used as vertices in the final graph, used for keyword extraction. Next, a graph-based ranking model is applied to assign a significance score to each topic and keywords are generated by selecting a candidate from each of the top-ranked topics. Network-based methodology has also been successfully applied to the task of topic extraction \cite{spitz2018entity}.

The method that we propose in this paper, RaKUn, is a graph-based keyword extraction method. We exploit some of the ideas from the area of graph aggregation-based learning, where, for example, graph convolutional neural networks and similar approaches were shown to yield high quality vertex representations by aggregating their neighborhoods' feature space \cite{cai2018comprehensive}. This work implements some of the similar ideas (albeit not in a neural network setting), where redundant information is aggregated into meta vertices 
in a similar manner. Similar efforts were shown as useful for hierarchical subnetwork aggregation in sensor networks \cite{chan2006secure} and in biological use cases of simulation of large proteins \cite{doruker2002dynamics}.

The main contributions of this paper are as follows. The notion of load centrality was to our knowledge not yet sufficiently exploited for keyword extraction. We show that this fast measure offers competitive performance to other widely used centralities, such as for example the PageRank centrality (used in \cite{mihalcea2004textrank}).
To our knowledge, this work is the first to introduce the notion of meta vertices with the aim of aggregating similar vertices, following similar ideas to the statistical method YAKE \cite{campos2018yake}, which is considered a state-of-the-art for the keyword extraction. Next, as part of the proposed RaKUn algorithm we extend the extraction from unigrams also to  bigram and threegram keywords based on load centrality scores computed for considered tokens. Last but not least, we demonstrate how arbitrary textual corpora can be transformed into weighted graphs whilst maintaining \emph{global} sequential information, offering the opportunity to exploit potential context not naturally present in statistical methods. 

The paper is structured as follows. We first present the text to graph transformation approach (Section~\ref{sec:rep}), followed by the introduction of the RaKUn keyword extractor (Section~\ref{sec:det}). 
We continue with qualitative evaluation 
(Section~\ref{sec:viz}) and quantitative evaluation (Section~\ref{sec:res}), 
before concluding the paper in Section 6.

\section{Transforming texts to graphs}
\label{sec:rep}
We first discuss how the texts 
are transformed to graphs, on which RaKUn operates. Next, we formally state the problem of keyword extraction and discuss its relation to graph centrality metrics.

\subsection{Representing text}
In this work we consider \emph{directed} graphs. Let $G = (V,E)$ represent a graph comprised of a set of vertices $V$ and a set of edges ($E \subseteq V \times V$), which are ordered pairs. Further, each edge can have a real-valued weight assigned.
Let $\mathcal{D}$ represent a document comprised of tokens $\{t_1,\dots,t_n\}$. The order in which tokens in text  
appear is known, thus $\mathcal{D}$ is a totally ordered set.
A potential way of constructing a graph from a document is by simply observing word co-occurrences. When two words co-occur, they are used as an edge. However, such approaches do not take into account the sequence nature of the words, meaning that the order is lost. We attempt to take this aspect into account as follows.
The given corpus is traversed, and for each element $t_i$, its successor $t_{i+1}$, together with a given element, forms a directed edge $(t_i,t_{i+1}) \in E$. Finally, such edges are weighted according to the number of times they appear in a given corpus. Thus the graph, constructed after traversing a given corpus, consists of all local neighborhoods (order one), merged into a single joint structure. Global contextual information is potentially kept intact (via weights), even though it needs to be detected via network analysis as proposed next.

\subsection{Improving graph quality by meta vertex construction}
A na\"ive approach to constructing a graph, as discussed in the previous section, commonly yields noisy graphs, rendering learning tasks harder. Therefore, we next discuss the selected approaches we employ in order to reduce both the computational complexity and the spatial complexity of constructing the graph, as well as increasing its quality (for the given down-stream task).

First, we consider the following heuristics which reduce the complexity of the graph that we construct for keyword extraction: Considered token length (while traversing the document $\mathcal{D}$, only tokens of length $\mu > \mu_{\textrm{min}}$ are considered), and next, lemmatization (tokens can be lemmatized, offering spatial benefits and avoiding redundant vertices in the final graph).  
The two modifications yield a potentially ``simpler'' graph,
which is more suitable and faster for mining.

Even if the optional lemmatization step is applied, one can still aim at further reducing the graph complexity by merging similar vertices. This step is called \textit{meta vertex construction}. The motivation can be explained by the fact, that even similar lemmas can be mapped to the same keyword (e.g., mechanic and mechanical;  normal and abnormal). This step also captures spelling errors (similar vertices that will not be handled by lemmatization), spelling differences (e.g., British vs. American English), non-standard writing (e.g., in Twitter data), mistakes in lemmatization or unavailable or omitted  lemmatization step.

\begin{figure}[ht]
\centering
\includegraphics[width=0.5\linewidth]{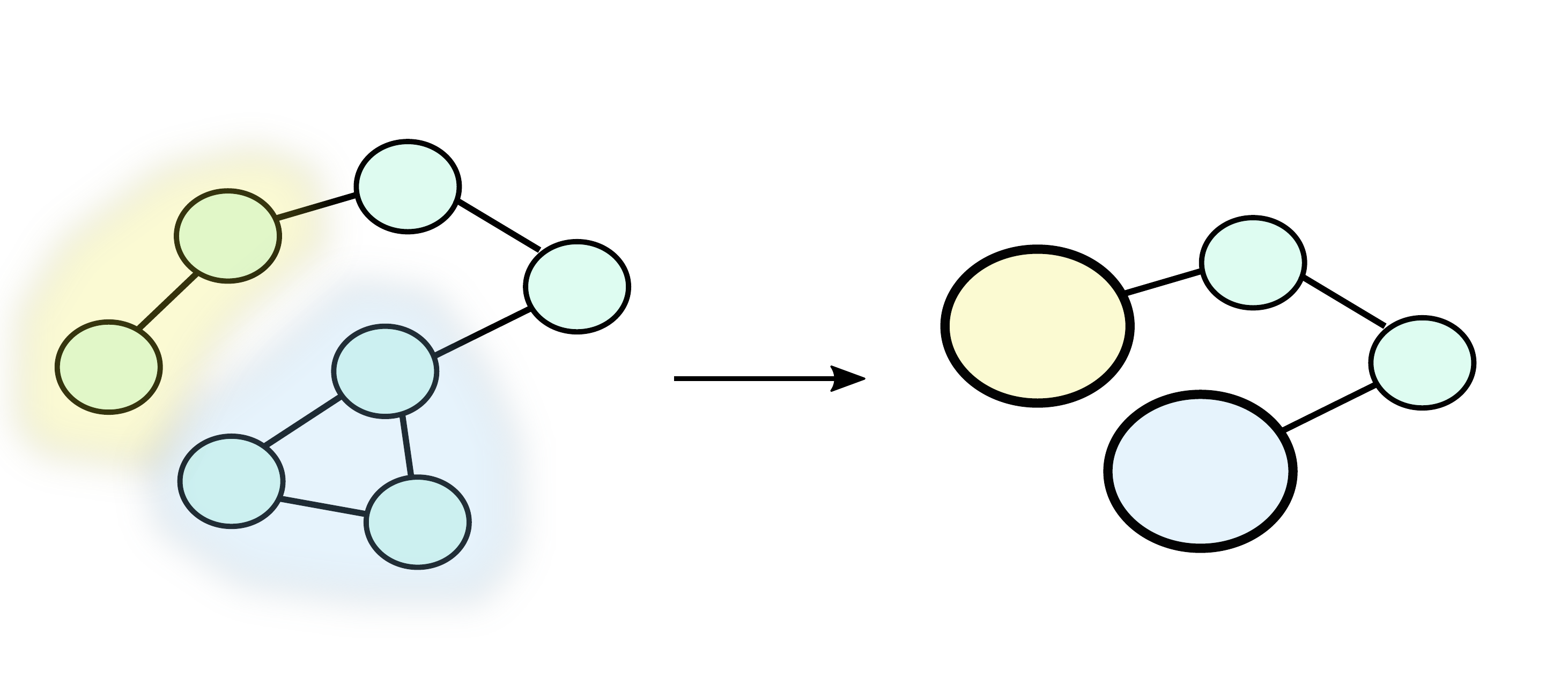}
\caption{Meta vertex construction. Sets of highlighted vertices are merged into a single vertex. The resulting graph has less vertices, as well as edges.}
\label{fig:hyp}
\end{figure}

The meta-vertex construction step works as follows. Let $V$ represent the set of vertices, as defined above. A meta vertex $M$ is comprised of a set of vertices that are elements of $V$, i.e. $M \subseteq V$. Let $M_i$ denote the $i$-th meta vertex. We construct a given $M_i$ so that for each $u \in M_i$, $u$'s initial edges (prior to merging it into a meta vertex) are rewired to the newly added $M_i$. Note that such edges connect to vertices which are not a part of $M_i$. Thus, both the number of vertices, as well as edges get reduced substantially. This feature is implemented via the following procedure:
\begin{enumerate}
\item Meta vertex candidate identification. Edit distance and word lengths distance are used to determine whether two words should be merged into a meta vertex (only if length distance threshold is met, the more expensive edit distance is computed). 
\item The meta vertex creation. As common identifiers, we use the stemmed version of the original vertices and if there is more than one resulting stem, we select the vertex from the identified candidates that has the highest centrality value in the graph and its stemmed version is introduced as a novel vertex (meta vertex). 

\item The edges of the words entailed in the meta vertex are next rewired to the meta vertex.
\item The two original words are removed from the graph.
\item The procedure is repeated for all candidate pairs.
\end{enumerate}
A schematic representation of meta vertex construction is shown in Figure~\ref{fig:hyp}. The yellow and blue groups of vertices both form a meta vertex, the resulting (right) graph is thus substantially reduced, both with respect to the number of vertices, as well as the number of edges.

\section{Keyword identification}
\label{sec:det}
Up to this point, we discussed how the graph used for keyword extraction is constructed. In this work, we exploit the notion of load centrality, a fast measure for estimating the importance of vertices in graphs. This metric can be defined as follows.

\subsubsection{Load centrality}
The load centrality of a vertex falls under the family of centralities which are defined based on the number of shortest paths that pass through a given vertex $v$, i.e. $c(v) =\sum_{t \in V}\sum_{s \in V} \frac{\sigma(s, t|v)}{\sigma(s, t)};t \neq s$, where $\sigma(s,t|v)$ represents the number of shortest paths that pass from vertex $s$ to vertex $t$ via $v$ and $\sigma(s,t)$ the number of all shortest paths between $s$ and $t$ (see \cite{brandes2008variants,PhysRevLett.87.278701}). The considered load centrality measure is subtly different from the better known betweenness centrality; specifically, it is assumed that each vertex sends a package to each other vertex to which it is connected, with routing based on a priority system: given an input of flow $x$ arriving at vertex $v$ with destination $v$', $v$ divides $x$ equally among all neighbors of minimum shortest path to the target. The total flow passing through a given $v$ via this process is defined as $v$'s load. Load centrality thus maps from the set of vertices $V$ to real values. 
For detailed description and computational complexity analysis, see \cite{brandes2008variants}. Intuitively, vertices of the graph with the highest load centrality represent key vertices in a given network. In this work, we assume such vertices are good descriptors of the input document (i.e. keywords). Thus, ranking the vertices yields a priority list of (potential) keywords. 

\subsubsection{Formulating the RaKUn algorithm}
We next discuss how the considered centrality is used as part of the whole keyword extraction algorithm RaKUn, summarized in Algorithm~\ref{algo:rakun}.
\begin{algorithm}[t]
\KwData{Document $D$, consisting of $n$ tokens $t_1,\dots,t_n$}
\Parameter{General: number of keywords $k$, minimal token length $\mu$; Meta vertex parameters: edit distance threshold $\alpha$, word length difference threshold $l$, Multi-word keywords parameters: path length $p$, 2-gram frequency threshold $f$}
\KwResult{A set of keywords $\mathcal{K}$}
\emph{corpusGraph}$\leftarrow$ EmptyGraph\Comment*[r]{Initialization.}
\For{$t_i \in D$}{
    \emph{edge} $\leftarrow (t_i,t_{i+1})$\;
    \If{\textit{edge} not in \textit{corpusGraph} and $\textrm{len}(t_i)$ $ \geq \mu$}{
        add \emph{edge} to \emph{corpusGraph} \Comment*[r]{Graph construction.}
    }
    updateEdgeWeight(\emph{corpusGraph}, \emph{edge})  \Comment*[r]{Weight update.}
}
\emph{corpusGraph} $\leftarrow$ generateMetaVertices(\emph{corpusGraph}, $\alpha$, $l$)\;
tokenRanks $\leftarrow$ loadCentrality(\emph{corpusGraph})  \Comment*[r]{Initial token ranks.}
scoredKeywords $\leftarrow$ generateKeywords($p$, $f$, tokenRanks)  \Comment*[r]{Keyword search.}
$\mathcal{K}$ = scoredKeywords[:$k$]\;
\Return{$\mathcal{K}$}
\caption{RaKUn algorithm.}
 \label{algo:rakun}
\end{algorithm}
The algorithm consists of three main steps described next. First, a graph is constructed from a given ordered set of tokens (e.g., a document) (lines 1 to 8). The resulting graph is commonly very sparse, as most of the words rarely co-occur. 
The result of this step is a smaller, denser graph, where both the number of vertices, as well as edges is lower. Once constructed, load centrality (line 10) is computed for each vertex. Note that at this point, should the top $k$ vertices by centrality be considered, only single term keywords emerge. As it can be seen from line 11, to extend the selection to 2- and 3-grams, the following procedure is proposed:
\begin{description}
\item[2-gram keywords.] Keywords comprised of two terms are constructed as follows. First, pairs of first order keywords (all tokens) are counted. If the support (= number of occurrences) is higher than $f$ (line 11 in Algorithm~\ref{algo:rakun}), the token pair is considered as potential 2-gram keyword. The load centralities of the two tokens are averaged, i.e. $c_{v} = \frac{c_1 + c_2}{2}$, and the obtained keywords are considered for final selection along with the computed ranks.
\item[3-gram keywords.] For construction of 3-gram keywords, we follow a similar idea to that of bigrams. The obtained 2-gram keywords (previous step) are further explored as follows. For each candidate 2-gram keyword, we consider two extension scenarios:
Extending the 2-gram from the left side. Here, the in-neighborhood of the left token is considered as a potential extension to a given keyword. Ranks of such candidates are computed by averaging the centrality scores in the same manner as done for the 2-gram case.
Extending the 2-gram from the right side. The difference with the previous point is that all outgoing connections of the rightmost vertex are considered as potential extensions. The candidate keywords are ranked, as before, by averaging the load centralities, i.e. $c_{v} = \frac{1}{3}\sum_{i=1}^{3}c_i$.
\end{description}

Having obtained a set of (keyword, score) pairs, we finally sort the set according to the scores (descendingly), and take top $k$ keywords as the result. We next discuss the evaluation the proposed algorithm.

\section{Qualitative evaluation}
\label{sec:viz}
\begin{figure}[ht]
\centering
\includegraphics[width=0.5\linewidth]{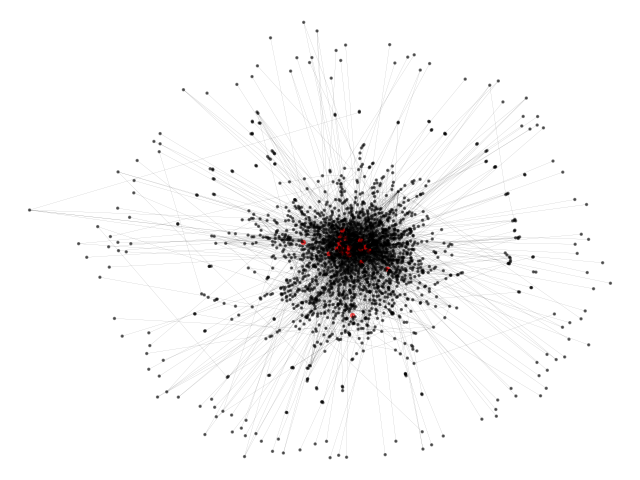}
\caption{Keyword visualization. Red dots represent keywords, other dots represent the remainder of the corpus graph.}
\label{fig:viz}
\end{figure}
\begin{figure}[ht]
\centering
\includegraphics[width=.7\linewidth]{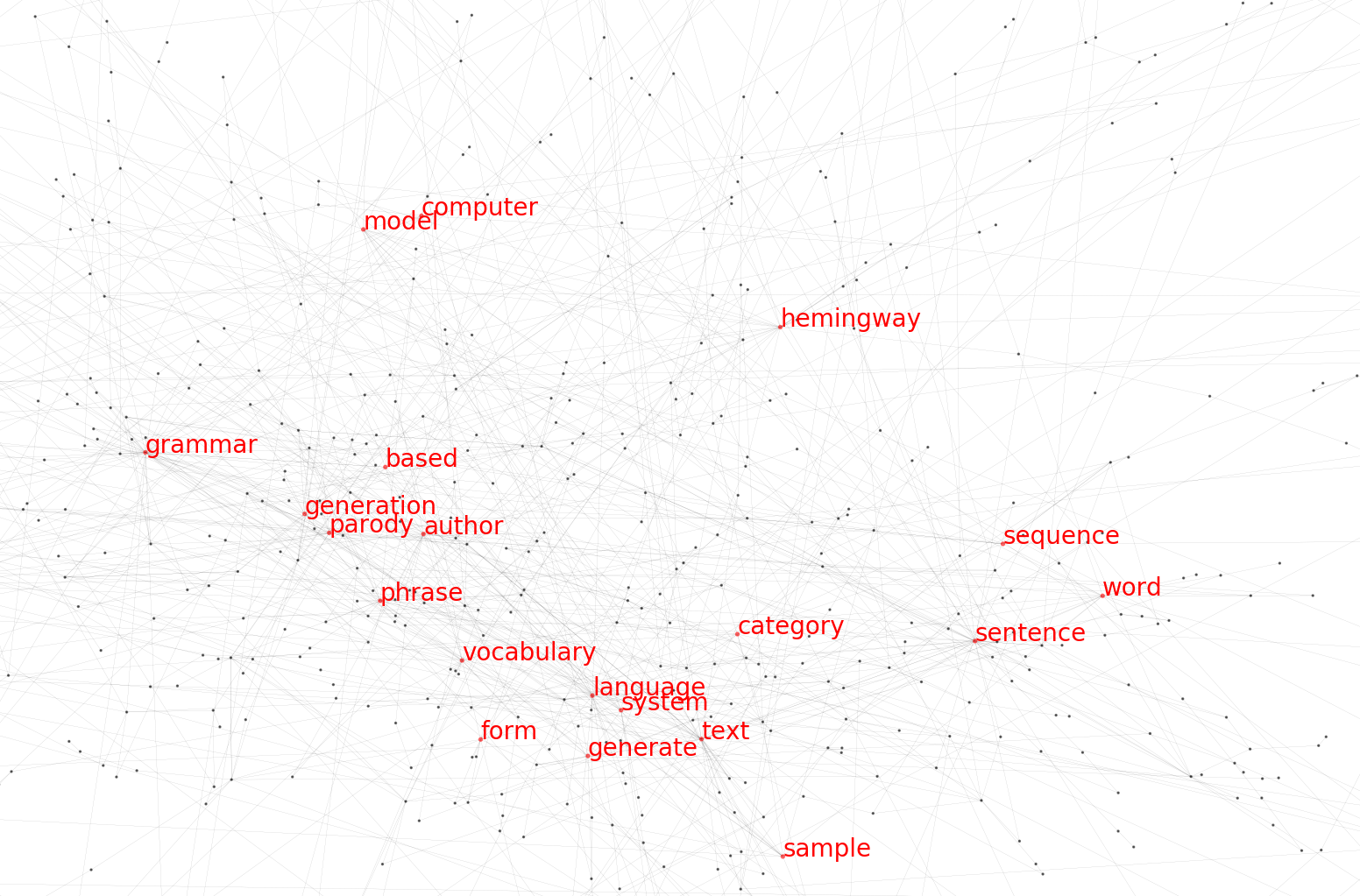}
\caption{Keyword visualization. A close-up view shows some examples of keywords and their location in the corpus graph. The keywords are mostly located in the central part of the graph.}
\label{fig:viz2}
\end{figure}

RaKUn can be used also for visualization of keywords in a given document or document corpus. A visualization of extracted keywords is applied to an example from wiki20 \cite{medelyan2008topic} (for dataset description see Section~\ref{sec:data}),
where we visualize both the global corpus graph, as well as a local (document) view where keywords are emphasized, see Figures~\ref{fig:viz} and~\ref{fig:viz2}, respectively. It can be observed that the global graph's topology is far from uniform --- even though we did not perform any tests of scale-freeness, we believe the constructed graphs are subject to distinct topologies, where keywords play prominent roles. 

\section{Quantitative evaluation}
\label{sec:res}
This section discusses the experimental setting used to validate the proposed RaKUn approach against state-of-the-art baselines. We first describe the datasets, and continue with the presentation of the experimental setting and results.

\subsection{Datasets}
\label{sec:data}
For RaKUn evaluation, we used 14 gold standard datasets from the list of \cite{campos2018yake,campos2}, from which we selected datasets in English. Detailed dataset descriptions and statistics can be found in Table \ref{tbl:q}, while full statistics and files for download can be found online\footnote{https://github.com/LIAAD/KeywordExtractor-Datasets}. Most datasets are from the domain of computer science or contain multiple domains. They are very diverse in terms of the number of documents---ranging from \textit{wiki20} with 20 documents to \textit{Inspec} with 2{,}000 documents, in terms of the average number of gold standard keywords per document---from 5.07 in \textit{kdd} to 48.92 in \textit{500N-KPCrowd-v1.1}---and in terms of the average length of the documents---from 75.97 in \textit{kdd} to \textit{SemEval2017} with 8332.34.


\ifx 
\begin{itemize}
\item \textit{500N-KPCrowd-v1.1} \cite{DBLP:journals/corr/MarujoVN13}, consisting of broadcast news transcriptions.
\item \textit{Inspec} \cite{Hulth:2003:IAK:1119355.1119383}, consisting of scientific journal papers from Computer Science collected between the years 1998 and 2002.
\item \textit{Nguyen2007} \cite{10.1007/978-3-540-77094-7_41}, consisting of 209 scientific conference papers.
\item \textit{PubMed}, consisting of full-text papers collected from PubMed Central.
\item \textit{Schutz2008} \cite{schutz2008keyphrase}, another dataset consisting of full-text papers collected from PubMed Central.
\item \textit{SemEval2010} \cite{Kim:2010:STA:1859664.1859668}, consisting of scientific papers extracted from the ACM Digital Library.
\item \textit{SemEval2017} \cite{DBLP:journals/corr/AugensteinDRVM17}, consisting of 500 paragraphs selected from 500 ScienceDirect journal articles, evenly distributed among the domains of Computer Science, Material Sciences and Physics.
\item \textit{citeulike180} \cite{Medelyan:2009:HTU:1699648.1699678}, consisting of full-text papers from the CiteULike.org platform.
\item \textit{fao30} \cite{doi:10.1002/asi.20790}, agricultural documents obtained from the two datasets based on Food and Agriculture Organization (FAO) of the United Nations.
\item \textit{fao780} \cite{doi:10.1002/asi.20790}, another dataset consisting of agricultural documents obtained from the two datasets based on Food and Agriculture Organization (FAO) of the United Nations. 
\item \textit{kdd} \cite{gollapalli2014extracting}, consisting of abstracts of papers collected from the ACM Conference on Knowledge Discovery and Data Mining (KDD) published during the period 2004-2014.
\item \textit{theses100}, consisting of 100 full master and Ph.D. theses from the University of Waikato, New Zealand.
\item \textit{wiki20} \cite{medelyan2008topic}, consisting of 20 English technical research reports covering different aspects of computer science.
\item \textit{www} \cite{gollapalli2014extracting}, consisting of abstracts of papers collected from the World Wide Web Conference (WWW) published during the period 2004-2014.
\end{itemize}
\fi

\begin{table}[h]
\centering
\caption{Selection of keyword extraction datasets in English language \vspace{0.3cm}}
\resizebox{0.99\textwidth}{!}{
\begin{tabular}{ |p{4cm}|p{8cm}||c|c|c|c|c}
 \hline
 Dataset &Desc.& No. docs & Avg. keywords & Avg. doc length\\
 \hline
500N-KPCrowd-v1.1 \cite{DBLP:journals/corr/MarujoVN13}& Broadcast news transcriptions & 500 & 48.92 & 408.33\\
Inspec \cite{Hulth:2003:IAK:1119355.1119383}  &Scientific journal papers from Computer Science collected between 1998 and 2002& 2000 & 14.62 & 128.20\\
Nguyen2007 \cite{10.1007/978-3-540-77094-7_41} & Scientific conference papers & 209 & 11.33 & 5201.09\\
PubMed & Full-text papers collected from PubMed Central& 500& 15.24 & 3992.78\\
Schutz2008\cite{schutz2008keyphrase} &Full-text papers collected from PubMed Central & 1231 & 44.69 & 3901.31\\
SemEval2010 \cite{Kim:2010:STA:1859664.1859668} & Scientific papers from the ACM Digital Library & 243 & 16.47 & 8332.34\\
SemEval2017 \cite{DBLP:journals/corr/AugensteinDRVM17} & 500 paragraphs selected from 500 ScienceDirect journal articles, evenly distributed among the domains of Computer Science, Material Sciences and Physics & 500 & 18.19 & 178.22\\
citeulike180 \cite{Medelyan:2009:HTU:1699648.1699678}&Full-text papers from the CiteULike.org & 180 & 18.42 & 4796.08\\
fao30 \cite{doi:10.1002/asi.20790} & Agricultural documents from two datasets based on Food and Agriculture Organization (FAO) of the UN& 30 & 33.23 & 4777.70\\
fao780 \cite{doi:10.1002/asi.20790} & Agricultural documents from two datasets based on Food and Agriculture Organization (FAO) of the UN  & 779 & 8.97 & 4971.79\\
kdd \cite{gollapalli2014extracting} & Abstracts from the ACM Conference on Knowledge Discovery and Data Mining (KDD) during 2004-2014 & 755 & 5.07 & 75.97\\
theses100  &Full master and Ph.D. theses from the University of Waikato& 100 & 7.67 & 4728.86\\
wiki20 \cite{medelyan2008topic}& Computer science technical research reports  & 20 & 36.50 & 6177.65\\
www \cite{gollapalli2014extracting} & Abstracts of WWW conference papers from 2004-2014 & 1330 & 5.80 & 84.08\\

 \hline
\end{tabular}
}
 \label{tbl:q}
\end{table}
\ifx
\begin{table}[]
\centering
\begin{tabular}{ |p{3cm}||c|c|c|c|c|}
 \hline
 Dataset & lang & No. docs & Avg. keywords & Avg. doc length\\
 \hline
500N-KPCrowd-v1.1 & en & 500 & 48.92 & 408.33\\
Inspec & en & 2000 & 14.62 & 128.20 & \\
Nguyen2007 & en & 209 & 11.33 & 5201.09 & \\
PubMed & en & 500 & 15.24 & 3992.78 & \\
Schutz2008 & en & 1231 & 44.69 & 3901.31 & \\
SemEval2010 & en & 243 & 16.47 & 8332.34 & \\
SemEval2017 & en & 500 & 18.19 & 178.22 & \\
citeulike180 & en & 180 & 18.42 & 4796.08 & \\
fao30 & en & 30 & 33.23 & 4777.70 & \\
fao780 & en & 779 & 8.97 & 4971.79 & \\
kdd & en & 755 & 5.07 & 75.97 & \\
theses100 & en & 100 & 7.67 & 4728.86 & \\
wiki20 & en & 20 & 36.50 & 6177.65 & \\
www & en & 1330 & 5.80 & 84.08 & \\
 \hline
\end{tabular}
 \label{tbl:q}
\end{table}
\fi

\subsection{Experimental setting}
We adopted the same evaluation procedure as used for the series of results recently introduced by YAKE authors \cite{campos2}\footnote{We attempted to reproduce YAKE evaluation procedure based on their experimental setup description and also thank the authors for additional explanation regarding the evaluation. For comparison of results we refer to their online repository https://github.com/LIAAD/yake \cite{campos2018yake}}. Five fold cross validation was used to determine the overall performance, for which we measured Precision, Recall and F1 score, with the latter being reported in Table \ref{tbl:pr}.\footnote{The complete results and the code are available at \url{https://github.com/SkBlaz/rakun}} Keywords were stemmed prior to evaluation.\footnote{This being a standard procedure, as suggested by the authors of YAKE.} As the number of keywords in the gold standard document is not equal to the number of extracted keywords (in our experiments $k$=10), in the recall we divide the correctly extracted keywords by the number of gold standard keywords.

\textbf{Selecting default configuration.} First, we used a dedicated run for determining the default parameters. The cross validation was performed as follows. For each train-test dataset split, we kept the documents in the test fold intact, whilst performing a grid search on the train part to find the best parametrization. Finally, the selected configuration was used to extract keywords on the unseen test set. For each train-test split, we thus obtained the number of true and false positives, as well as true and false negatives, which were summed up and, after all folds were considered, used to obtain final F1 scores, which served for default parameter selection. The grid search was conducted over the following parameter range
Num keywords: $10$, Num tokens (the number of tokens a keyword can consist of): Count threshold (minimum support used to determine potential bigram candidates): 
Word length difference threshold (maximum difference in word length used to determine whether a given pair of words shall be aggregated): $[0,2,4]$,  Edit length difference (maximum edit distance allowed to consider a given pair of words for aggregation): $[2,3]$, Lemmatization: [yes, no].

Even if one can use the described grid-search fine-tunning procedure to select the best setting for individual datasets, we observed that in nearly all the cases the best settings were the same. We therefore selected it as the \emph{default}, which can be used also on new unlabeled data. The default parameter setting was as follows. The number of tokens was set to 1, Count threshold was thus not needed (only unigrams), for meta vertex construction Word length difference threshold was set to 3 and Edit distance to 2. Words were initially lemmatized. Next, we report the results using these selected parameters (same across all datasets), by which we also test the general usefulness of the approach. \footnote{Note that at the time of writing of this paper,  repository of YAKE! did not include exact evaluation code, thus we attempted to reproduce the results to the best of our knowledge. Recently published full version of YAKE! implies the computation of recall differs, thus RaKUn's results are not directly comparable.}

\subsection{Results}
The results are presented in Table \ref{tbl:pr}, where we report on F1 with the default parameter setting of RaKUn, together with the results from related work, as reported in the github table of the YAKE \cite{campos2018yake}\footnote{https://github.com/LIAAD/yake/blob/master/docs/YAKEvsBaselines.jpg (accessed on: June 11, 2019)}.
\begin{table}[t]
\centering
\caption{Performance comparison with state-of-the-art approaches.}
\vspace{0.3cm}
\resizebox{\textwidth}{!}{
\begin{tabular}{l|p{1.2cm}p{1.3cm}p{1.2cm}p{1.2cm}p{1.2cm}p{1.2cm}p{1.2cm}p{1.2cm}p{1.5cm}}
 \hline
 Dataset & RaKUn  & YAKE & Single Rank & KEA & KP-MINER & Text Rank & Topic Rank & Topical PageRank\\
 \hline
 \rowcolor{green!8} 500N-KPCrowd-v1.1 &  \textbf{0.428}&  0.173 & 0.157 & 0.159 & 0.093 & 0.111 & 0.172 & 0.158\\
Inspec &  0.054&  0.316 & \textbf{0.378} & 0.150 & 0.047 & 0.098 & 0.289 & 0.361\\
Nguyen2007 &  0.096 &  0.256 & 0.158 & 0.221 & \textbf{0.314} & 0.167 & 0.173 & 0.148\\
 PubMed &  0.075 & 0.106 & 0.039 & \textbf{0.216} & 0.114 & 0.071 & 0.085 & 0.052\\
 \rowcolor{green!8} Schutz2008 & \textbf{0.418} &  0.196 & 0.086 & 0.182 & 0.230 & 0.118 & 0.258 & 0.123\\
 SemEval2010 &  0.091 & 0.211 & 0.129 & 0.215 & \textbf{0.261} & 0.149 & 0.195 & 0.125\\
SemEval2017 & 0.112  & 0.329 & \textbf{0.449} & 0.201 & 0.071 & 0.125 & 0.332 & 0.443\\
citeulike180 & 0.250 &  0.256 & 0.066 & \textbf{0.317} & 0.240 & 0.112 & 0.156 & 0.072\\
 \rowcolor{green!8} fao30& \textbf{0.233}  & 0.184 & 0.066 & 0.139 & 0.183 & 0.077 & 0.154 & 0.107\\
 fao780 & 0.094 & \textbf{0.187} & 0.085 & 0.114 & 0.174 & 0.083 & 0.137 & 0.108\\
kdd & 0.046 &  \textbf{0.156} & 0.085 & 0.063 & 0.036 & 0.050 & 0.055 & 0.089\\
theses100 &  0.069 & 0.111 & 0.060 & 0.104 & \textbf{0.158} & 0.058 & 0.114 & 0.083\\
 \rowcolor{green!8} wiki20 &  \textbf{0.190} & 0.162 & 0.038 & 0.134 & 0.156 & 0.074 & 0.106 & 0.059\\
www & 0.060 & \textbf{0.172} & 0.097 & 0.072 & 0.037 & 0.059 & 0.067 & 0.101 \\
\end{tabular}
 \label{tbl:pr}
}
\end{table}
We first observe that on the selection of datasets, the proposed RaKUn wins more than any other method. We also see that it performs notably better on some of the datasets, whereas on the remainder it performs worse than state-of-the-art approaches. Such results demonstrate that the proposed method finds keywords differently, indicating load centrality, combined with meta vertices, represents a promising research venue. The datasets, where the proposed method outperforms the current state-of-the-art results are: \textit{500N-KPCrowd-v1.1}, \textit{Schutz2008}, \textit{fao30} and \textit{wiki20}. In addition, RaKUn also achieves competitive results on \textit{citeulike180}. A look at the gold standard keywords in these datasets reveals that they contain many single-word units which is why the default configuration (which returns unigrams only) was able to perform so well.

Four of these five datasets (\textit{500N-KPCrowd-v1.1}, \textit{Schutz2008}, \textit{fao30}, \textit{wiki20}) are also the ones with the highest average number of keywords per document  with at least 33.23 keywords per document, while the fifth dataset (\textit{citeulike180}) also has a relatively large value (18.42). 
Similarly, four of the five well-performing datasets (\textit{Schutz2008}, \textit{fao30}, \textit{citeulike180}, \textit{wiki20}) include long documents (more than 3,900 words), with the exception being \textit{500N-KPCrowd-v1.1}. For details, see Table \ref{tbl:q}. We observe that the proposed RaKUn outperforms the majority of other competitive graph-based methods. For example, the most similar variants Topical PageRank and TextRank do not perform as well on the majority of the considered datasets. Furthermore, RaKUn also outperforms KEA, a supervised keyword learner (e.g., very high difference in performance on 500N-KPCrowd-v1.1 and Schutz2008 datasets), indicating unsupervised learning from the graph's structure offers a more robust keyword extraction method than learning a classifier directly.

\section{Conclusions and further work}
In this work we proposed RaKUn, a novel unsupervised keyword extraction algorithm which exploits the efficient computation of load centrality, combined with the introduction of meta vertices, which notably reduce corpus graph sizes. The method is fast, and performs well compared to state-of-the-art such as YAKE and graph-based keyword extractors. In further work, we will test the method on other languages. We also believe additional semantic background knowledge information could be used to prune the graph's structure even further, and potentially introduce keywords that are inherently not even present in the text (cf.\cite{vskrlj2019towards}). The proposed method does not attempt to exploit meso-scale graph structure, such as convex skeletons or communities, which are known to play prominent roles in real-world networks and could allow for vertex aggregation based on additional graph properties.
We believe the proposed method could also be extended using the Ricci-Oliver \cite{jin2007discrete} flows on weighted graphs.

\subsubsection{Acknowledgements}
The work was supported by the Slovenian Research Agency through a young researcher grant [B\v{S}], core research programme (P2-0103), and projects Semantic Data Mining for Linked Open Data (N2-0078) and Terminology and knowledge frames across languages (J6-9372). This work was supported also by the EU Horizon 2020 research and innovation programme, Grant No. 825153, EMBEDDIA (Cross-Lingual Embeddings for Less-Represented Languages in European News Media). The results of this publication reflect only the authors’ views and the EC is not responsible for any use that may be made of the information it contains. We also thank the authors of YAKE for their clarifications.

\bibliographystyle{splncs04}
\bibliography{references}

\end{document}